\documentclass[10pt,twocolumn,letterpaper]{article}

\usepackage{acpr}
\usepackage{times}
\usepackage{epsfig}
\usepackage{graphicx}
\usepackage{amsmath}
\usepackage{amssymb}
\usepackage{booktabs} 
\usepackage{multirow}
\usepackage{bigstrut} 
\graphicspath{{./figures/}}

\usepackage[pagebackref=true,breaklinks=true,letterpaper=true,colorlinks,bookmarks=false]{hyperref}

\acprfinalcopy % *** Uncomment this line for the final submission

\ifacprfinal\fi
\begin{document}

%%%%%%%%% TITLE
\title{GM-Net: Learning Features with More Efficiency}

\author{Yujia Chen\\
School of Computer \& Communication \\ Engineering\\ 
University of Science and Technology Beijing\\
{\tt\small cyj.issac@gmail.com}
\and
Ce Li\\
China University of Mining \\\& Technology, Beijing \\
University of Chinese Academy of Sciences\\
{\tt\small licekong@gmail.com}
}

\maketitle
%\thispagestyle{empty}

%%%%%%%%% ABSTRACT
\begin{abstract}
Deep Convolutional Neural Networks (CNNs) are capable of learning unprecedentedly effective features from images. Some researchers have struggled to enhance the parameters' efficiency using grouped convolution. However, the relation between the optimal number of convolutional groups and the recognition performance remains an open problem. In this paper, we propose a series of Basic Units (BUs) and a two-level merging strategy to construct deep CNNs, referred to as a joint Grouped Merging Net (GM-Net), which can produce joint grouped and reused deep features while maintaining the feature discriminability for classification tasks. Our GM-Net architectures with the proposed BU\_A (dense connection) and BU\_B (straight mapping) lead to significant reduction in the number of network parameters and obtain performance improvement in image classification tasks. Extensive experiments are conducted to validate the superior performance of the GM-Net than the state-of-the-arts on the benchmark datasets, e.g., MNIST, CIFAR-10, CIFAR-100 and SVHN.
\end{abstract}

%%%%%%%%% BODY TEXT
\section{Introduction}\label{sec:intro}

Convolution Neural Networks (CNNs) has drawn lots of attention recently due to strong power of extracting structural and semantic features. %It has almost become a necessary in computer vision tasks like image classification\cite{srivastava2014jmlr,he2015deep,huang2016deepnet,huang2016arxiv}, object detection\cite{girshick2015iccv,lin2016arxiv}, and face detection\cite{hu2016arxiv}
Effective learning strategies enable CNNs, even a shallow one, to represent any bounded polynomial under certain conditions\cite{andoni2014icml}, which has brought remarkable performance improvement in vision tasks.

%Making a CNN deeper is one of the attempts to get higher level semantic information. 
After LeNet\cite{lecun1998gradient-based} demonstrated the effectiveness (5 layers), the performance of Alexnet\cite{krizhevsky2012nips} (8 layers) and VGGnet\cite{russakovsky2015imagenet} (19 layers) show that deeper is better. ResNet\cite{he2015deep} and Highway Networks\cite{srivastava2015training} have made it to more than 100 layers. However, deeper model is not necessarily better than shallower one. Information may lose sharply as going back through the network, thus layers in the beginning seldom get optimized, and these redundant layers will eventually become a hindrance.%\cite{lee2015taistats}. He \etal\cite{he2016eccv} tried to tackle this problem with identity mapping, which allows gradients to flow directly to any layer. %But Veit \etal\cite{veit2016nips} testified that, the valid length of paths that contribute to gradient is no longer than half of the longest path in ResNet, much shorter than expected. 
Wu \etal\cite{wu2016wider} introduced a shallower but wider model that gain better results without going very deep. Yet, it is not a once-for-all solution because it still has too much parameters that influence the speed and overfitting risk. 
\begin{figure}[t]
	\begin{center}
		\includegraphics[width=0.7\linewidth]{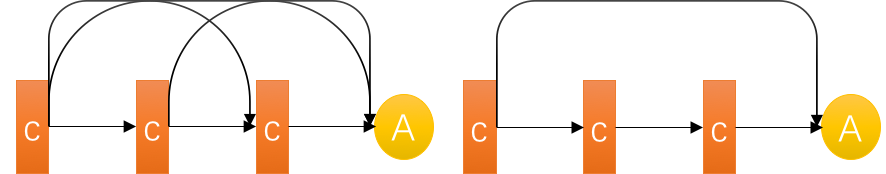}
	\end{center}
	\vspace{-6pt}
	\caption{{\bf Left:} BU\_A with densely connections, where C stands for convolution unit and A stands for adaption unit. {\bf Right:} BU\_B with straight mapping.}
	\vspace{-10pt}
	\label{fig:buab}
\end{figure}

\begin{figure*}[t]
	\begin{center}
		\includegraphics[width=0.67\linewidth]{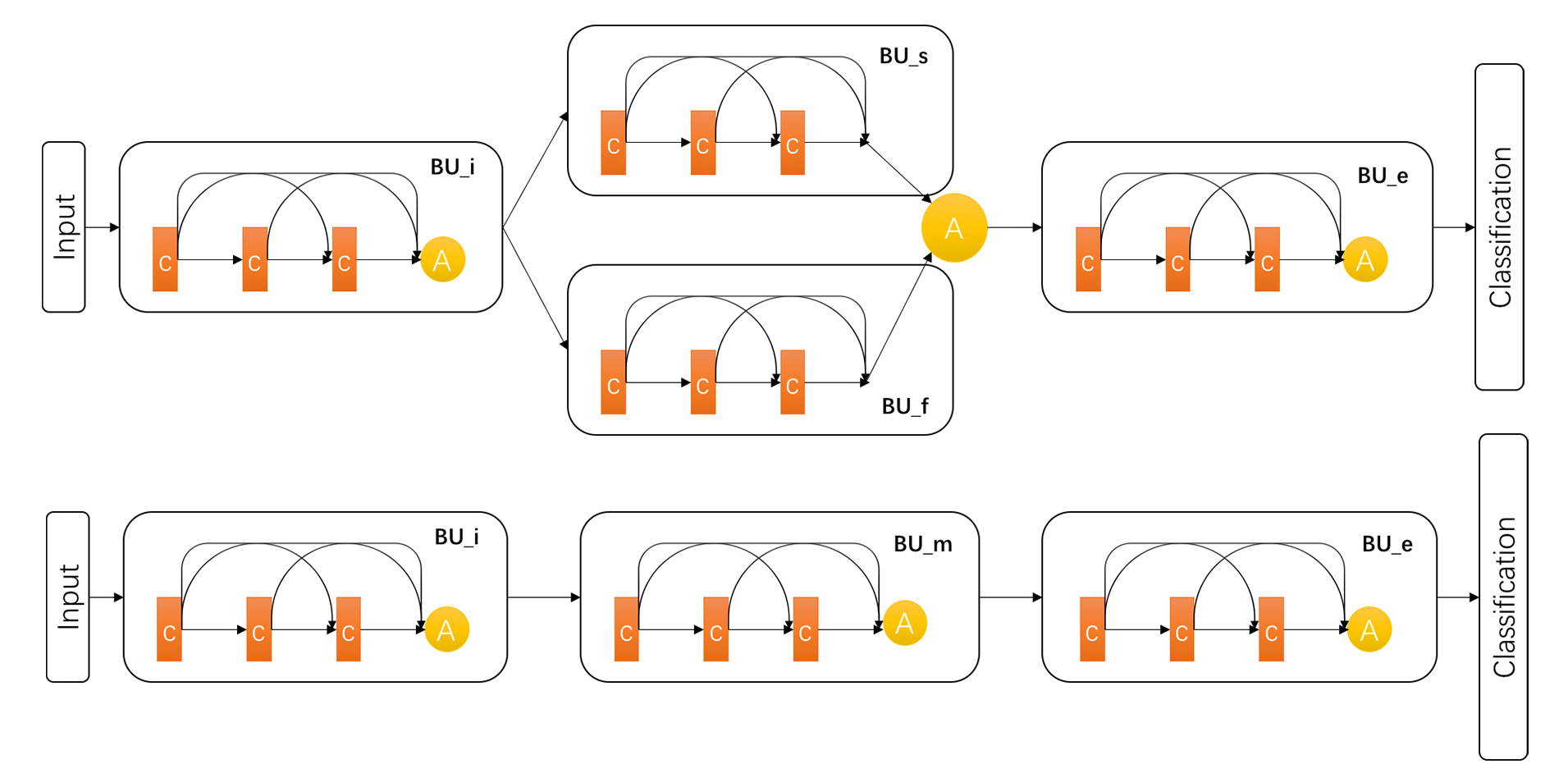}
	\end{center}
	\vspace{-6pt}
	\caption{Our proposed structures. {\bf Top:} GM-Net with BU\_A. {\bf Bottom:} Baseline model with BU\_A.}
	\vspace{-10pt}
	\label{fig:ourapp}
\end{figure*}

Although grouped convolution\cite{krizhevsky2012nips} was first used in Alexnet to compensate for the hardware deficiency, its utilization of making the connection sparse leading to more useful feature is verified by many scholars. Besides, we find that an appropriate merging strategy is very important while applying grouped convolution contiguously, because each group has limited learning power and non-negligible information loss. Furthermore, outputs with different convolution groups focus on various contents, so that the combination of different convolution group numbers will produce more powerful feature maps.

In this paper, we propose an architecture with well-designed basic units (BUs) (Figure \ref{fig:buab}), slightly different with basic form, detailed in Section \ref{sec:approach}. As shown in Figure \ref{sec:approach}, we define four units as BU\_i (input), BU\_f (full), BU\_s (single) and BU\_e (end). After BU\_i, the input will flow through BU\_f and  BU\_s separately in order to get representations with different structural and semantic information. Then, these two flows are merged and fed into BU\_e, leading to capture relations between channels. Particularly, our model aims to extract better feature via deliberately-designed feature reuse instead of pursuing being deeper. In addition, to reduce abundant parameters, we apply grouped convolution in almost each convolutional layer. 

BU, composed of 3 convolution units and 1 adaption unit, is the crux of our models. Two kinds of BUs are proposed as BU\_A and BU\_B (Figure \ref{fig:buab}). Enlightened by\cite{huang2016arxiv}, BU\_A applies densely connection, which connects each layer to others. However, it originally aims to alleviate gradient vanishing, strengthen feature propagation and reduce parameter number with concatenation operation. We take this advantage mainly by replacing concatenation with summation operation, since simply stacking different layers together can't learn features with high-level information sufficiently, demonstrated in Section \ref{sec:experiments}. Moreover, the negative influence of more parameters with summation operation is mitigated by grouped convolution. Different with BU\_A, the BU\_B applies straight mapping from beginning to the end. We observe that two BUs have similar performance and BU\_A is slightly better on more complex datasets (Table \ref{tab:errorrate}), and evaluate two models on four benchmarks: MNIST\cite{le1998mnist}, CIFAR-10\cite{krizhevsky2009techport}, CIFAR-100\cite{krizhevsky2009techport} and SVHN\cite{netzer2011reading}. With only 29 layers and 2 paths, they gains competitive results with state-of-the-arts, which demonstrates that the extracted feature are informative enough to represent the contents, even if our model contains far less parameters. 

The paper is organized as follows. Section \ref{sec:relatedwork} reviews related works. Section \ref{sec:approach} describes our proposed model. Experiments are in Section \ref{sec:experiments} and conclusion in Section \ref{sec:conclusion}.
%-------------------------------------------------------------------------
\section{Related Work}\label{sec:relatedwork}
{\bf Overfitting:} Overfitting has long been a major issue in computer vision tasks, since millions of parameters tend to remember each training data\cite{zhang2016arxiv}. Dropout\cite{srivastava2014jmlr} provides an effective regularization that prevents co-adaptions on training data by randomly dropping out neurons. It can be explained as an ensemble of sparse neural networks to reduce overfitting risks. Similarly, stochastic depth\cite{huang2016deepnet} and drop path\cite{larsson2016fractalnet} utilize the idea of only training part of the network to optimize models. We attempt to make parameters sparse and compact by dropping connections with special basic units. Sparse models reduce overfitting generally, but underfitting will occur if it is too sparse. This problem is handled in our models with a unique merging strategy.

\begin{table*}[t]
  \renewcommand{\arraystretch}{1.15}
	\vspace{-6pt}
	\centering
	\begin{tabular}{|c|c|c|c|}
		\hline
		BU    & Output & BU\_A (B) & Baseline \\
		\hline\hline
		\multirow{3}[6]{*}{BU\_i} & \multirow{3}[6]{*}{$16 \times 16$} & $3 \times 3 \times 64$     &  $3 \times 3 \times 64$  \\
		\cline{3-4}          &       & $\left[ {\begin{array}{*{20}{c}}
			1 \times 1 \times 64\\
			3 \times 3 \times 128
			\end{array}} \right] \times 3, g=8 $     & $\left[ {\begin{array}{*{20}{c}}
			1 \times 1 \times 64\\
			3 \times 3 \times 128
			\end{array}} \right] \times 3, g=8 $ \bigstrut\\
		\cline{3-4}          &       & $\begin{array}{*{20}{c}}
		{3 \times 3 \times 128,g = 4}\\
		{2 \times 2 \rm{~AvePool,~stride2}}
		\end{array}$     & $\begin{array}{*{20}{c}}
		{3 \times 3 \times 128,g = 4}\\
		{2 \times 2 \rm{~AvePool,~stride2}}
		\end{array}$ \\
		\hline
		BU\_f & $16 \times 16$ &  $\left[ {\begin{array}{*{20}{c}}
			1 \times 1 \times 128\\
			3 \times 3 \times 256
			\end{array}} \right] \times 3, g=1 $     & \multirow{4}[8]{*}{-} \\
		\cline{1-3}    \multirow{2}[4]{*}{BU\_s} & \multirow{2}[4]{*}{$16 \times 16$} &  $1 \times 1 \times 256, g=1$     &  \\
		\cline{3-3}          &       &   $3 \times 3 \times 256,\times 3, g=256$      &  \\
		\cline{1-3}    AU    &  $8 \times 8$     &    $\begin{array}{*{20}{c}}
		{3 \times 3 \times 256,g = 4}\\
		{2 \times 2\rm{~AvePool,~stride2}}
		\end{array}$    &  \\
		\hline
		\multirow{2}[4]{*}{BU\_m} & \multirow{2}[4]{*}{$8 \times 8$} & \multirow{2}[4]{*}{-} & $\left[ {\begin{array}{*{20}{c}}
			1 \times 1 \times 128\\
			3 \times 3 \times 256
			\end{array}} \right] \times 3, g=8 $  \\
		\cline{4-4}          &       &       & $\begin{array}{*{20}{c}}
		{3 \times 3 \times 256,g = 4}\\
		{2 \times 2\rm{~AvePool,~stride2}}
		\end{array}$ \\
		\hline
		\multirow{2}[4]{*}{BU\_e} & \multirow{2}[4]{*}{$8 \times 8$} &    $\left[ {\begin{array}{*{20}{c}}
			1 \times 1 \times 192\\
			3 \times 3 \times 384
			\end{array}} \right] \times 3, g=8 $   & $\left[ {\begin{array}{*{20}{c}}
			1 \times 1 \times 192\\
			3 \times 3 \times 384
			\end{array}} \right] \times 3, g=8 $ \\
		\cline{3-4}          &       &     $1 \times 1 \times 384,g = 4$  & $1 \times 1 \times 384,g = 4$ \\
		\hline
		& $1 \times 1$      & \rm{Global AvePool, 10d-fc, softmax} & \rm{Global AvePool, 10d-fc, softmax} \\
		\hline
		\#Params &  -  & $1.5$\rm M  & $0.7$ \rm M \\
		\hline
	\end{tabular}%
	\vspace{4pt}
	\caption{GM-Net and baseline model for CIFAR-10 and SVHN datasets, where $g$ denotes the number of groups and AU denotes adaption unit.}
	\vspace{-10pt}
	\label{tab:architecture}%
\end{table*}%
{\bf Feature Reuse:} Combining low-level features that contain detailed information and high-level features that contain semantic information together has demonstrated its practicability in recent works. Residual representation and identity mapping in ResNet\cite{he2016eccv} value in both forward and backward propagation, and residual learning is a good reformulation or preconditioning that simplify optimization and fully take advantages of all the knowledges flowing in forward. And identity mapping provides a simpler method. DenseNet\cite{huang2016arxiv} uses different level features by concatenating thin feature maps, which also reduce parameters dramatically. All methods suggest that finding an effective way of feature reuse will boost model's accuracy.

{\bf Grouped Convolution:} Ascribe to the computing limitation, AlexNet\cite{krizhevsky2012nips} has to split its feature maps into two groups, which enlightens scholars nowadays to find high-efficient architectures. Grouped convolution hasn't got much attention until ResNeXt\cite{xie2016arxiv} pointed out the significance of cardinality, explained as the number of convolutional groups. By increasing cardinality, ResNeXt\cite{xie2016arxiv} increases its accuracy when maintaining model's complexity and parameter numbers. Xception\cite{chollet2016arxiv} proposed depth-wise separable convolution, or channel-wise convolution, where each channel is deemed as a convolution group. As achieving higher performance, an open question is still left behind: What is the optimal number of convolutional groups?

%-------------------------------------------------------------------------
\section{Approach}\label{sec:approach}
\subsection{Architecture Overview}
As shown in Figure \ref{sec:approach}, images are first fed into input basic unit (denoted as BU\_i), and then flow through full-convolution unit (denoted as BU\_f) and single-channel convolution unit (denoted as BU\_s) separately. Finally, with end-flow unit (denoted as BU\_e) merging two paths together, output is computed on high-level semantic features. Our baseline model only replaces BU\_f and BU\_s with a middle unit (denoted as BU\_m), which has the same settings with BU\_e. Table \ref{tab:architecture} describes the two proposed models.

Generally, GM-Net contains two paths in the middle. After splitting channels into different groups, three convolution units are performed contiguously. In Caffe\cite{jia2014caffe}, grouped convolution is realized by matching input groups with output groups in order, so it can also be regarded as each single group flows through three convolution units uninterruptedly. In this way, each BU has paths of the same number with group numbers. The method also selects the best local feature of the same location with different filters. Similarly, we realize it by squeezing the main flow into some groups and apply apt merging strategy. Besides, sparse connection caused by grouped convolution ensures that no redundant information is learned. 

Merging is another key of our model. Limited by the channel numbers of each group, the learning power has its upper bound. Consequently, contiguous grouped convolution can't be performed in a long sequence. That's why we merge all paths with an adaption unit after only three convolution units. More details are presented in Section \ref{sec:mergingstrategy}.

BU\_A and BU\_B is also designed specifically to reinforce feature reuse. BU\_A contains densely connected mapping to reuse features from all previous stages, reducing the information loss incurred by small number of parameters. And BU\_B with a straight mapping follows the idea in\cite{he2016eccv}, enabling each unit to learn in a sequence. Also, both of two methods provides effective means in backward propagation.

\begin{table*}[t]
	\vspace{-6pt}
	\centering
	\begin{tabular}{|l|c|c|c|c|c|c|c|c|}
		\hline
		Method & \multicolumn{1}{c|}{Depth} & \#Params & \multicolumn{1}{c|}{C10} & \multicolumn{1}{c|}{C10+} & \multicolumn{1}{c|}{C100} & \multicolumn{1}{c|}{C100+} & \multicolumn{1}{c|}{SVHN} & \multicolumn{1}{c|}{MNIST} \\
		\hline\hline
		Network in Network\cite{lin2016iclr} & \multicolumn{1}{c|}{-} & -     & 10.41 & 8.81  & 35.68 & \multicolumn{1}{c|}{-} & 2.35  & 0.45\\
		%\hline
		DropConnet\cite{wan2013icml} & \multicolumn{1}{c|}{-} & -     & \multicolumn{1}{c|}{-} & 9.32  & \multicolumn{1}{c|}{-} & \multicolumn{1}{c|}{-} & 1.94  & 0.23 \\
		%\hline
		%Deep Supervised Net\cite{lee2015taistats} & \multicolumn{1}{l|}{-} & -     & 9.78  & 8.22  & \multicolumn{1}{l|}{-} & 34.57 & 1.92  & 0.39 \\
		\hline
		Highway Network\cite{srivastava2015training} & 19    & 2.3M  & \multicolumn{1}{c|}{-} & 7.72  & \multicolumn{1}{c|}{-} & 32.39 & \multicolumn{1}{c|}{-} & 0.45 \\
		%\hline
		ResNet\cite{he2015deep} & 110   & 1.7M  & \multicolumn{1}{l|}{-} & 6.61  & \multicolumn{1}{c|}{-} & \multicolumn{1}{c|}{-} & \multicolumn{1}{c|}{-} & \multicolumn{1}{c|}{-} \\
		%\hline
		Stochastic Depth\cite{huang2016deepnet} & 110   & 1.7M  & 11.66 & 5.23  & 37.8  & 24.58 & 1.75  & \multicolumn{1}{c|}{-} \\
		%\hline
		Fractal Net\cite{larsson2016fractalnet} & 21    & 38.6M & 10.18 & 5.22  & 35.34 & 23.3  & 2.01  & \multicolumn{1}{c|}{-} \\
		%\hline
		Wide ResNet\cite{zagoruyko2016wideresnet} & 16    & 8.9M  & \multicolumn{1}{c|}{-} & 4.97  & \multicolumn{1}{c|}{-} & 22.89 & \multicolumn{1}{c|}{-} & \multicolumn{1}{c|}{-} \\
		%\hline
		Fitnet4-LSUV\cite{xie2017arxiv} & 17    & 2.5M  & \multicolumn{1}{l|}{-} & 6.06  & \multicolumn{1}{c|}{-} & 27.66 & \multicolumn{1}{c|}{-} & \multicolumn{1}{c|}{-} \\
		%\hline
		Tree+Max-Avg\cite{lee2016icais} & \multicolumn{1}{c|}{-} & -     & 7.62  & 6.05  & 32.37 & \multicolumn{1}{c|}{-} & 1.69  & 0.31\\
		%\hline
		DenseNet (growth rate =12)\cite{huang2016arxiv} & 40    & 1M    & 7     & 5.24  & 27.55 & 24.42 & 1.79  & \multicolumn{1}{c|}{-} \\
		\hline
		Baseline & 23    & 0.7M  & 7.13  & 5.98  & 27.59 & 25.81 & 2.39  & 0.3 \\
		%\hline
		{\bf BU\_A} & {\bf 29}    & {\bf 1.5M}  & {\bf 6.81}  & {\bf 5.15}  & {\bf 27.09} & {\bf 24.97} & {\bf 1.88}  & {\bf 0.21} \\
		%\hline
		{\bf BU\_B} &{\bf 29}    & {\bf 1.5M}  & {\bf 6.77}  & {\bf 5.62}  & {\bf 27.08} & {\bf 25.22} & {\bf 1.92}  & {\bf 0.22} \\
		\hline
	\end{tabular}%
	\vspace{4pt}
	\caption{Error rates on CIFAR and SVHN datasets. Models that contains too many parameters are not presented in this table. Our models obtain comparable results even with far fewer parameters compared to many state-of the-arts.}
	\vspace{-10pt}
	\label{tab:errorrate}%
\end{table*}%
\subsection{Basic Unit Design}\label{sec:basicunitdesign}
Convolution unit is composed of a bottleneck where a $1 \times 1$ convolution is applied before a $3 \times 3$ convolution. In convolution unit. Adaption unit consists of a pooling layer with stride 2 after a $3 \times 3$ or $1 \times 1$ convolution layer. To obtain different representative features, we use standard convolution in BU\_f and channel-wise grouped convolution in BU\_s. We empirically demonstrated the effectiveness of this method in Section \ref{sec:experimentonarchi}. The number of parameters of a grouped convolution can be computed as:
\begin{equation}
k \times k \times \frac{c}{g} \times \frac{o}{g} \times g
\label{eq:groupconv}
\end{equation}
where $k$ denotes kernel size, $c$ denotes the number of input channels, $o$ denotes the number of output channels, and $g$ is the group number. In this way, parameters are $g$ times smaller, and more compact features can be learned.

\subsection{Merging Strategy}\label{sec:mergingstrategy}
Merging is applied densely in our model to compensate  
for the disadvantages of grouped convolution. Our merging strategy has two levels. In operation level, $1 \times 1$ convolution inside a convolution unit has two functions: reducing channels and merging information. So we don't use grouped convolution on it. Thus, merging is applied inside every convolution unit. In unit level, BU\_A and BU\_B merges features from different layers to reinforce detailed information, demonstrated in Section \ref{sec:experimentonarchi}.

\subsection{Implementation Details}
For each convolution unit, $1 \times 1$ convolution reduce the number of channels with the factor of 0.5. All convolution operation is composed of conv-BN-ReLU. We use 8 groups for $3 \times 3$ convolution and 4 groups for adaption unit. Besides, dropout is used in BU\_i and BU\_e to reduce overfitting with keep probability of 0.8. In experiment settings, we use Nesterov optimization with momentum 0.9. Base learning rate is 0.1 or 0.01 (MNIST) and is reduced 10 times after half and 3/4 of total epochs. SVHN and MNIST datasets are trained 200 epochs and 300 epochs for other datasets.
%-------------------------------------------------------------------------
\section{Experiments}\label{sec:experiments}
In this section, we first conduct experiments on MNIST\cite{le1998mnist}, CIFAR-10\cite{krizhevsky2009techport}, CIFAR-100\cite{krizhevsky2009techport} and SVHN\cite{netzer2011reading} datasets in Section \ref{sec:experimentondata}. We also conduct ablation experiments, without data augmentation or other tricks, to further analyze the practicability of our model with CIFAR-10 dataset in Section \ref{sec:experimentonarchi}.

\subsection{Experiments on Different Datasets}\label{sec:experimentondata}
{\bf MNIST.} It contains 60000 hand-written digits for training and 10000 for testing. The images are collected from 250 writers and have shapes of $1 \times 28 \times 28$ where 1 denotes one channel. 

{\bf CIFAR.} It consists of 60000 $32 \times 32$ colored natural scene images, 10000 of which are used for testing. Specifically, CIFAR-10 (C10) contains 10 classes, with 5000 training images and 1000 testing images per class, while CIFAR-100 (C100) has 100 classes, 500 training images and 100 testing images for each classes. Channel means are computed and subtracted in preprocessing. We also apply standard augmentation\cite{srivastava2015training,huang2016deepnet,larsson2016fractalnet,zagoruyko2016wideresnet,huang2016arxiv,xie2017arxiv} (marked as C10+ and C100+), leading into higher accuracy. 

{\bf SVHN.} It is a real-world dataset obtained from house numbers in Google Street View images. It consists 10 classes, where 73257 digits for training, 26032 digits for testing, and 531131 additional. All digits have been resized to 32-by-32 pixels. The task is to classify the central digit into a correct class. %We didn't use additional dataset for training.

{\bf Result Analysis.} In our method, the efficiency of parameters is the key. We observe that even without large number of parameters, our models obtain accuracy that are comparable with many state-of-the-art methods (Table \ref{tab:errorrate}). Our baseline model contains only 0.7M parameters, but performs almost the same as Fractal Net\cite{larsson2016fractalnet} which has {\bf 40 times} more parameters. This fully demonstrates that our models extract most useful information without overfitting. Moreover, with alike parameters, BU\_A and BU\_B outperform well-designed architectures like Highway Network\cite{srivastava2015training} and Fitnet4-LSUV\cite{xie2017arxiv} by a large margin on CIFAR-10 and CIFAR-100. We believe it is the better output representations that accounts for the result. 

On real-world SVHN dataset with more complex background, though our models obtain good result, there is still a lot space to improve on. For example, increasing param-eters can help grasp more crucial information. On MNIST dataset, GM-Net outperforms lots of well-designed archit-ectures and achieves state-of-the-art accuracy. Since it is a comparatively simpler dataset, useful information is easier to extract and utilize. We ascribe the results to the informative representation of features.

Besides, BU\_A and BU\_B perform similar, while B\_A is slightly better on more complex datasets like CIFAR-100 and SVHN. Because CIFAR-100 has fewer images for each class and SVHN contains many ambiguous digits, dense connections can act as good form of feature reuse to effectively compensate for information loss during forward propagation.

\subsection{Experiments on Architecture}\label{sec:experimentonarchi}
\subsubsection{Study on Grouped Convolution}
As original input images and bottom convolution layers contain highly correlated adjacent pixels, the reduction of connections in bottom layers can lead to sparse representation, forcing to abandon redundant information. Besides, making parameters compact on top layers may also lead to more representative features. We compare putting blocks with grouped convolution at different locations (Table \ref{tab:convgroup}). The result shows that model without grouped convolution has the lowest accuracy. As expected, setting grouped convolution on block1 gains an increase of 0.02\% than block3. The low image resolution of datasets may explain the little gap — there is less need to reduce correlation on images with low resolution. Last, if we split the channels into groups for both block1 and block3, accuracy will also increase by 0.64\%. Thanks to merging strategy, information is efficiently utilized with the breakthrough in the limitations of group's learning ability.

\begin{table}
	\vspace{-6pt}
	\centering
	\begin{tabular}{|l|c|}
		\hline
		Settings & \multicolumn{1}{l|}{Accuracy on CIFAR-10} \\
		\hline\hline
		No groups & 92.55\% \\
		%\hline
		Groups conv on block1 & 93.03\% \\
		%\hline\textsc{}
		Group conv on block3 & 93.01\% \\
		%\hline
		Both block1 and block3 ({\bf ours}) & {\bf 93.19\%} \\
		\hline
		8+2   & 92.86\% \\
		%\hline
		8+4 ({\bf ours}) & {\bf 92.87\% }\\
		%\hline
		8+4\&16+8 & 92.35\% \\
		%\hline
		4+4\&4+4 & 92.01\% \\
		\hline
	\end{tabular}%
	\vspace{4pt}
	\caption{Different settings on convolutional groups.}
	\vspace{-10pt}
	\label{tab:convgroup}%
\end{table}%

\begin{figure}
	\vspace{-3pt}
	\begin{center}
		\includegraphics[width=0.75\linewidth]{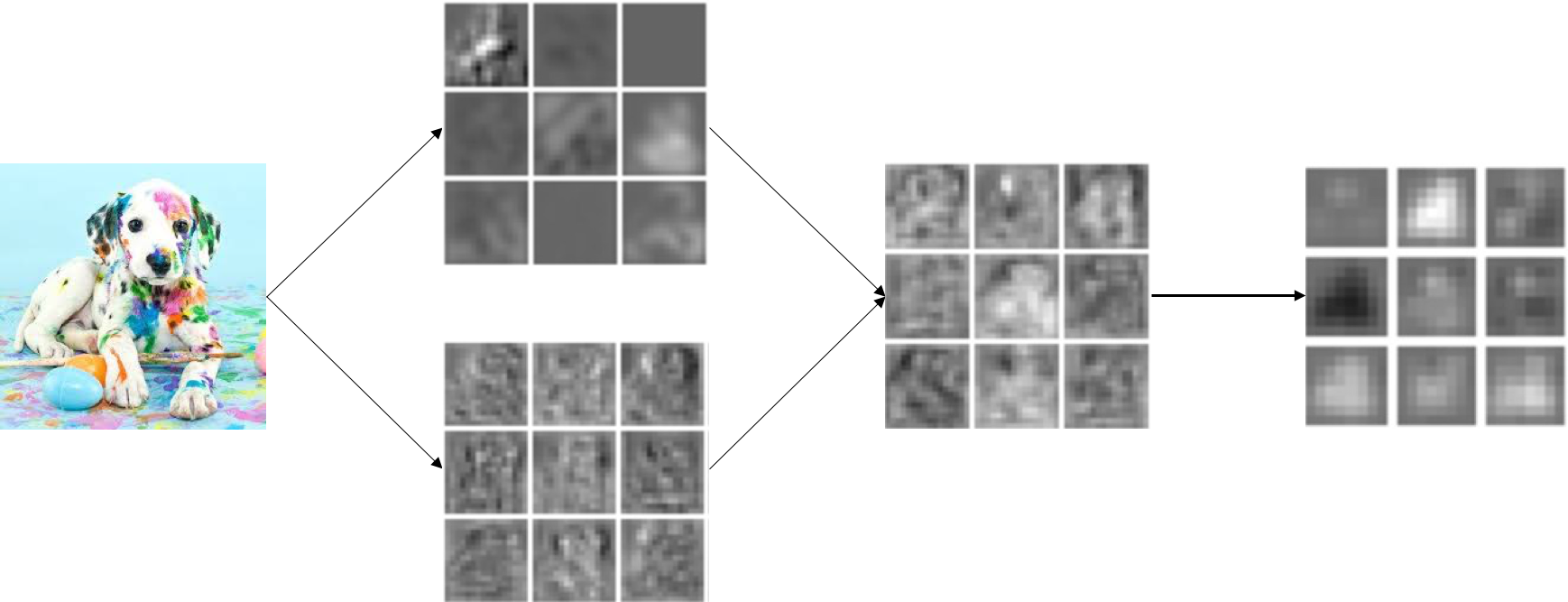}
	\end{center}
	\caption{Feature maps visualization. The features on the top are extracted from the end of BU\_s, the bottom ones are from the end of BU\_f, the third ones are from the adaption unit after BU\_s and BU\_f, and the last ones are from the last convolution layer. Features become more discriminable with the flow as expected.}
	\vspace{-6pt}
	\label{fig:visual}
\end{figure}

\begin{figure}[htbp]
	\begin{center}
		\includegraphics[width=0.75\linewidth]{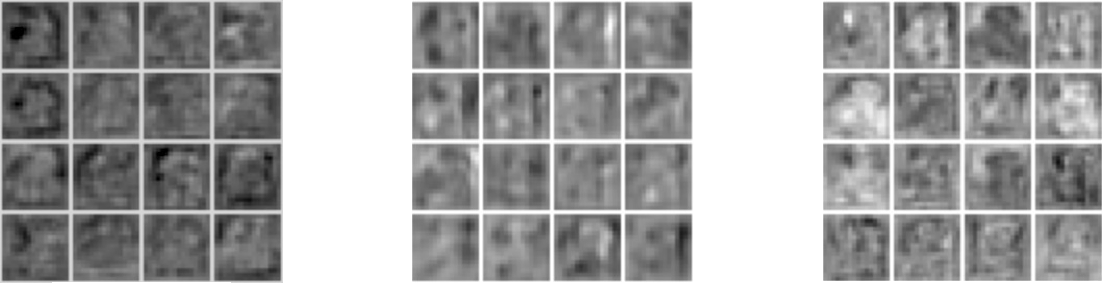}
	\end{center}
	\caption{Features with different merging strategy experimented with baseline model. Features are extracted from the adaption unit after BU\_m. {\bf Left:} operation-level merging is taken out, resulting in an accuracy drop by 0.75\% and darker feature maps. {\bf Middle:} unit-level merging is taken out, resulting in an accuracy drop by 0.4\% and blurred feature maps. {\bf Right:} features with our merging strategy, showing shaper objects and informative representation.}
	\vspace{-15pt}
	\label{fig:merge}
\end{figure}
How many groups should be used still remains an open question. We try two methods on baseline model — keeping group number and keeping channel numbers (Table \ref{tab:convgroup}). For the former one, we set 8 for group numbers of convolution units, and 2 and 4 for adaption unit in both BU\_i and BU\_m (marked as 8+2 or 8+4). For the latter method, channel number is fixed at 8 for each group in convolution unit and 16 in adaption unit (marked as 8+4 \& 16+8). A setting where 4 groups used in both convolution and adaption units is also presented as a comparison.

The fact that our setting outperforms others verifies the advantages of the merging strategy. The increase of channel number from the latter block blender flows together, grasping more information. If we decrease the group number in adaption unit to 2, performances are similar while number of parameters vary sharply. The comparison between the first setting and the last setting also bespeaks that our grouped convolution helped grasp useful features instead of causing information loss.

\subsubsection{Comparison on feature reuse method}
We can see from Table \ref{tab:errorrate} that GM-Net outperform our baseline by around 0.35-0.83\%, verifying that combining features that propagated from different convolution groups can effectively increase the accuracy. Figure \ref{fig:visual} and Figure \ref{fig:merge} show the function of our merging strategy. We can observe from Figure \ref{fig:visual} that features after BU\_s are sparser and concentrate only on a specific part on each feature map, while features after BU\_f contains a lot more redundant information. By combining them together after an adaption unit, we find the part of the object is more highlighted and the motif shows more discriminability, which leads to a more informative classification layer afterwards. 

We also conduct ablation experiments to see how merging strategy influence the features. While accuracy drops by 0.75\% on CIFAR-10, darker feature maps in Figure \ref{fig:merge} (a) shows the significant information loss without operation-level merging. If we only take out unit-level merging, result drops by 0.4\%. Feature maps is lighter but more blurred. This may match the learning limitation with groups. With our two-level yet simple merging strategy, feature maps become sharper and still informative enough. 

\begin{figure}
	\begin{center}
		\includegraphics[width=0.65\linewidth]{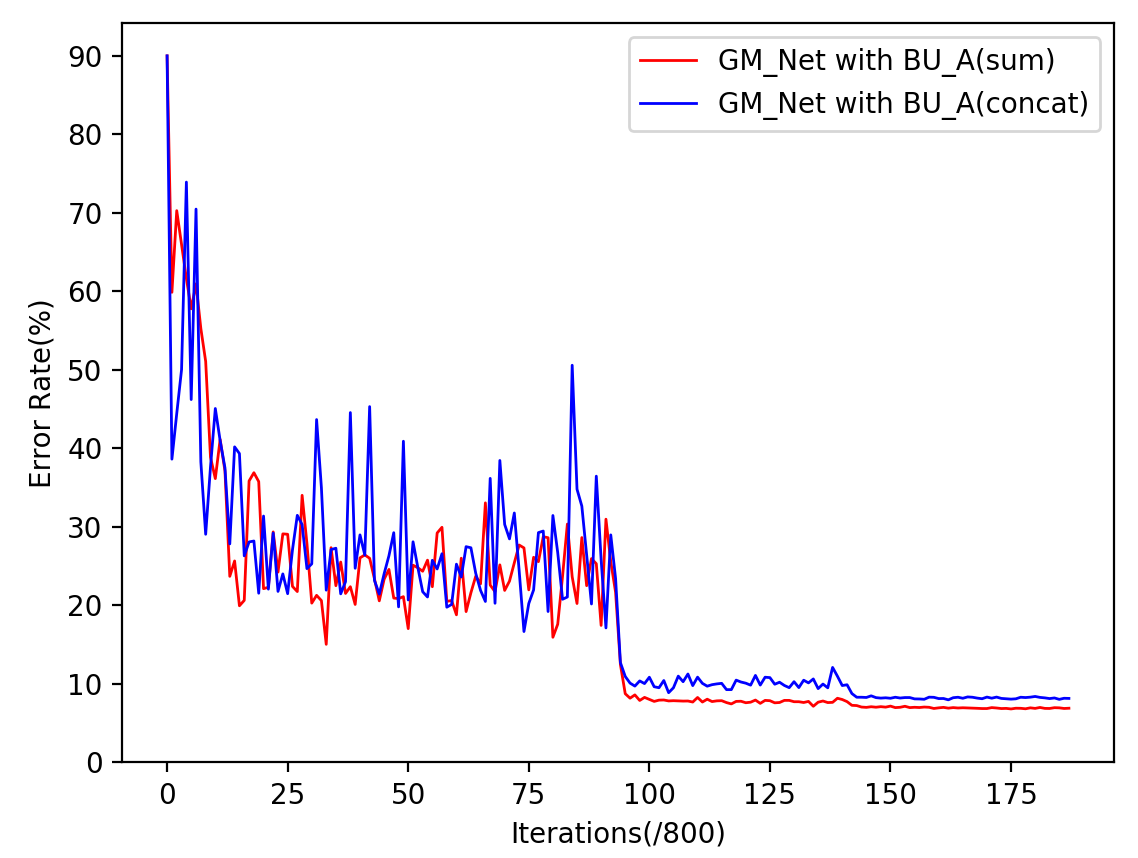}
	\end{center}
	\caption{CIFAR-10 test error rate of GM-Net with two kinds of merging methods.}
	\vspace{-15pt}
	\label{fig:testerror}
\end{figure}

Finally, we plot test error for model with sum merging and concatenate merging (Figure \ref{fig:testerror}) to verify superior performance of sum operation than concatenation. Accuracy drops by 1.1\% with the latter method. Specifically, the basic unit for concatenate merging is like \cite{huang2016arxiv} but shorter. To keep the parameters in a same level, we set growth rate as 20, 40, 60 for BU\_i, BU\_f and BU\_e. The number can be easily computed with Equation (\ref{eq:groupconv}). BU\_s is performed as described in Section \ref{sec:basicunitdesign}. Furthermore, no grouped convolution is used in the latter method, because it has already contained a similar idea of compacting parameters.

%-------------------------------------------------------------------------
\section{Conclusion}\label{sec:conclusion}
With the study on sparse issue of grouped convolution for applications in computer vision in this paper, we propose a joint Grouped Merging Network (GM-Net) with two basic unit forms, BU\_A (dense connection) and BU\_B (straight mapping), and a simple two-level merging strategy to compensate for the reduction of parameters and learn features with more efficiency. Joint grouped and reused semantic features are produced and demonstrates superior efficiency on utilizing parameters than the state-of-the-arts. The improvement of experimental result is obtained in four image classification tasks and ascribed to the better feature representation. In the future, we will keep making CNNs lighter and learning more representative features. 

%-------------------------------------------------------------------------
%-------------------------------------------------------------------------
%------------------------------------------------------------------------
%\section*{Acknowledgement}

{\small
\bibliographystyle{unsrt}%ieee
\bibliography{egbib}
}

\end{document}